\documentclass{article} % For LaTeX2e
\usepackage{iclr2024_conference,times}

\usepackage{amsmath,amsfonts,bm}

\def\eqref#1{equation~\ref{#1}}
\def\1{\bm{1}}

\DeclareMathAlphabet{\mathsfit}{\encodingdefault}{\sfdefault}{m}{sl}
\SetMathAlphabet{\mathsfit}{bold}{\encodingdefault}{\sfdefault}{bx}{n}

\usepackage{times}
\usepackage{latexsym}
\usepackage[colorlinks=true,citecolor=.]{hyperref}
\usepackage{url}
\usepackage{bbm}
\usepackage{amsmath,amsfonts,bm}
\usepackage{graphicx}
\usepackage{multirow}
\usepackage{makecell}
\usepackage{array}
\usepackage[para]{threeparttable}
\usepackage{booktabs}
\usepackage{tabularx}
\usepackage{xspace}
\usepackage{subcaption}
\usepackage{multirow}
\usepackage[normalem]{ulem}
\usepackage{longtable}
\usepackage{amsmath}
\usepackage{hyperref}

\title{\raisebox{-0.1cm}{\includegraphics[width=0.7cm, height=0.637cm]{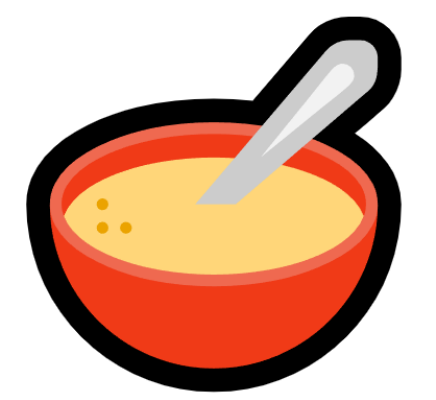}}Personalized Soups: Personalized Large \\Language Model Alignment via Post-hoc\\ Parameter Merging}

\author{Joel Jang{\textsuperscript{1,2}}\quad Seungone Kim {\textsuperscript{3}} \quad Bill Yuchen Lin{\textsuperscript{2}} \quad Yizhong Wang{\textsuperscript{1}} \quad Jack Hessel{\textsuperscript{2}} \\ \textbf{Luke Zettlemoyer{\textsuperscript{1}} \quad Hannaneh Hajishirzi{\textsuperscript{1,2}} \quad Yejin Choi{\textsuperscript{1,2}} \quad Prithviraj Ammanabrolu{\textsuperscript{4}}} \\
{\textsuperscript{1}}University of Washington \quad{\textsuperscript{2}}Allen Institute for AI \quad{\textsuperscript{3}}KAIST AI\quad{\textsuperscript{4}}UC San Diego  \\
\texttt{joeljang@cs.washington.edu} \\
}

\definecolor{salmon}{HTML}{F69289}

\iclrfinalcopy % Uncomment for camera-ready version, but NOT for submission.
\begin{document}

\maketitle

\begin{abstract}
While Reinforcement Learning from Human Feedback (RLHF) aligns Large Language Models (LLMs) with general, aggregate human preferences, it is suboptimal for learning diverse, individual perspectives. In this work, we study Reinforcement Learning from Personalized Human Feedback (RL$\mathcal{P}$HF) problem, wherein LLMs are aligned to multiple (sometimes conflicting) preferences by modeling alignment as a Multi-Objective Reinforcement Learning (MORL) problem. Compared to strong single-objective baselines, we show that we can achieve \textbf{personalized alignment} by decomposing preferences into multiple dimensions. These dimensions are defined based on personalizations that are declared as desirable by the user. In this work, we show that they can be efficiently trained independently in a distributed manner and combined effectively post-hoc through parameter merging.~\footnote{Code: \url{https://github.com/joeljang/RLPHF}}
\end{abstract}

\section{Introduction}
Reinforcement Learning from Human Feedback (RLHF)~\citep{Nakano2021WebGPTBQ, Ouyang2022TrainingLM, Bai2022TrainingAH, dubois2023alpacafarm, bai2022constitutional} typically optimizes a policy model that receives training signals from a single reward model that aims to capture the \textit{general} preferences of a population. In this work, we instead propose Reinforcement Learning from \textit{Personalized} Human Feedback (RL$\mathcal{P}$HF), a new, multi-objective formulation of the human preference alignment problem, where Large Language Models (LLMs) are trained to be efficiently aligned with a range of different, potentially personalized combinations of human preferences. 

We model RL$\mathcal{P}$HF as a Multi-Objective Reinforcement Learning (MORL) problem, which allows training the policy model with multiple, \textit{conflicting} objectives since it aims to vary the importance of each objective during inference. In existing RLHF formulations, pairwise human feedback is collected by asking human annotators to choose which model response is \textit{generally} better and is used to train a general reward model. This makes \textit{implicit} assumptions that may not hold for everyone. For example, recent work has shown that LLMs aligned with RLHF prefer verbose output generations~\citep{zheng2023judging, dubois2023alpacafarm, wang2023far, singhal2023long}. 
We aim to support a wider range of multifaceted preferences that are explicitly declared as desirable by the user---giving the user control over the facets of output text they want to see as well as the personal data they wish to reveal to the model.
We collect \textit{personalized} human feedback corresponding to multiple such dimensions, noting that they may also be conflicting in nature.

We first implement a strong MORL baseline called \textsc{Prompted-MORL} where there are multiple reward signals for each of the objectives (preferences) given via prompts during RL training. Next, we propose \textsc{Personalized Soups}, a method that circumvents simultaneously optimizing multiple preferences by first optimizing multiple policy models each with distinct preferences with Proximal Policy Optimization (PPO) and merging the parameters of the policy models whose preferences we want to composite together on the fly during inference. This \textit{modular} approach significantly reduces the computational complexity from exponential to linear in relation to the total number of unique preferences. Furthermore, since \textsc{Personalized Soups} does not have to be trained in a multitask fashion, it does not require \textit{re-training} the underlying policy every time a novel preference (objective) is added.

\begin{figure}[h] 
\centering
\includegraphics[width=\textwidth]{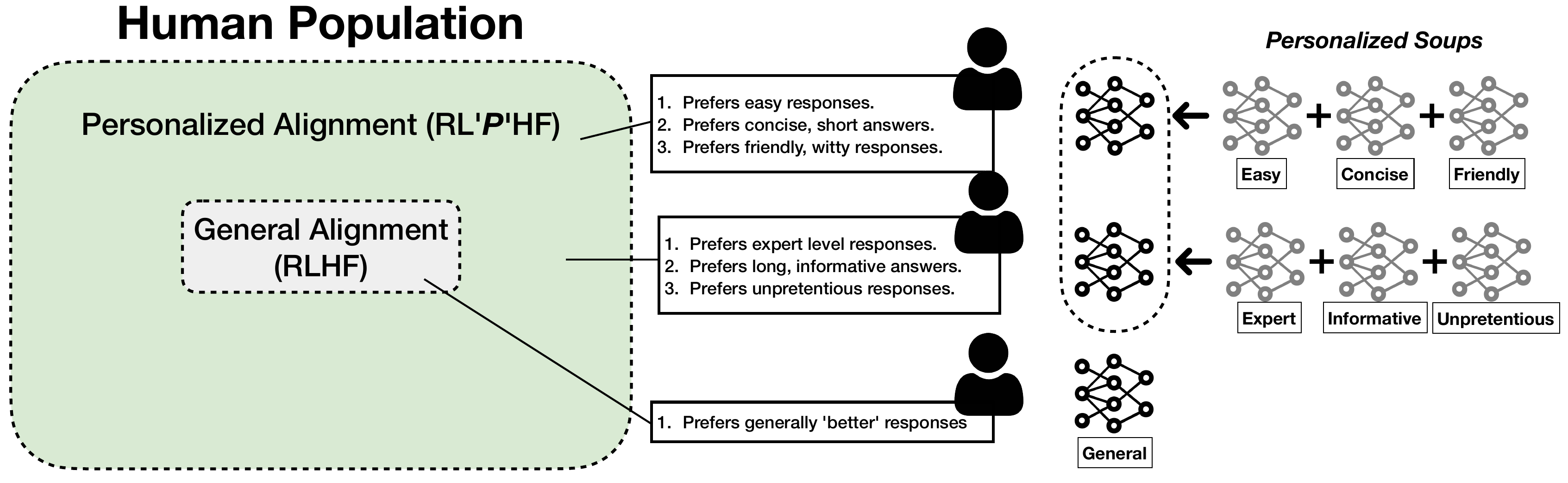}
\caption{Current RLHF only tackles general alignment while RL$\mathcal{P}$HF is able to take into account multifaceted human preferences, thus providing a more personalized form of model-to-human alignment.}
\label{fig:main}
\end{figure}

We empirically show that by transforming the problem of aligning LLMs to human preferences into a MORL problem, we are able to have \textbf{personalized alignment} that provides a deeper level of adaptation to individual users that supervised fine-tuning, RLHF, and prompting cannot attain. We also emphasize the modularity of \textsc{Personalized Soups} by performing experiments in a scenario where the user additionally writes novel preferences that they want to integrate with existing preferences. We show that in this scenario, \textsc{Personalized Soups} still performs competitively to \textsc{Prompted-MORL} while being exponentially more efficient through parameter merging.

\section{Related Work}
\paragraph{Aligning Language Models To Human Preferences}
Incorporating human preference feedback into a reward model, and subsequently optimizing a language model to output text that reward model scores highly with an RL algorithm, has been shown to result in language models that generate outputs humans generally prefer \citep{ouyang2022training}. This process has been applied to summarization~\citep{ziegler2019fine,stiennon2020learning,wu2021recursively}, answering questions with long-form answers using text retrieved from the web~\citep{nakano2021webgpt,menick2022teaching}, generating engaging responses in a dialogue settings~\citep{thoppilan2022lamda,cohen2022dynamic} and following human instructions~\citep{kojima2021continual,suhr2022continual,kim2023aligning}.

However, the standard RLHF setup commonly addressed in prior work assumes a reward model that accounts only for \emph{average} annotator preference, i.e., the fact that different users may desire different outputs, even for the same prompt, is ignored \citet{casper2023open}. Individual preferences can vary not only on aesthetic axes, but also on semantics. For example,  \citet{santurkar2023whose} use public opinion polling to show that ``default" LLM preferences vary in their degree of expressed-opinion alignment with different average opinions among demographic groups.\footnote{\citet{feng2023pretraining} suggests that ``default" LLM expressed opinions stem directly from the pretraining data.} % and show that the human feedback-tuned LMs have some inherent biases, while poorly reflecting some surface groups.
\citet{kirk2023personalisation} defines a taxonomy and policy framework for the alignment of LLMs with personalized feedback. While \citet{wu2023fine} performs fine-grained RLHF which is very similar in spirit and allows personalization, our work develops MORL algorithms for scenarios where there are \textit{conflicting} preferences, not only orthogonal objectives.

\paragraph{Multi-objective Reinforcement Learning (MORL)}
\label{subsec:morl}

In this work, we propose formulating LLM personalization as a MORL problem, which was typically studied in decision-making tasks~\citep{hayes2022practical} that aims to tackle the problem of simply optimizing by a single, scalar, additive reward function~\citep{sutton2018reinforcement}, which possesses many limitations such as (1) suboptimal solutions due to lack of representation~\citep{hayes2022practical}, (2) lack of explainability of distinct objectives, and (3) ensuring fair outcomes for multiple participants~\citep{vamplew2018human, siddique2020learning}.

Previous work has aimed to alleviate these problems through novel MORL methods~\citep{van2013scalarized, van2014multi, yang2019generalized, xu2020prediction}. Other work aims to solve complex problems such as water management, military purchasing, wind farm control, etc.~\citep{hayes2022practical} by converting the single-objective RL problem into a MORL problem. In this work, we convert the problem of aligning LLMs to human preferences into a MORL problem to (1) provide a more optimal solution for each individual, (2) allow users to dynamically choose the distinct objectives they want to optimize, and (3) ensure fairness by allowing preferences that may be in the long-tail to be integrated.

\paragraph{Personalization in Natural Language Processing}
Personalization in Natural Language Processing (NLP) has mainly been focused on creating personalized dialogue agents~\citep{zhang2018personalizing, mazare-etal-2018-training, zheng2019personalized, wu2021personalized, xu-etal-2022-beyond}, where the task is to create chitchat agents that are engaging with distinct personas based on user profile (e.g. gender, age, residence, etc.) or past user history data (e.g. Reddit posts, etc.). Another line of work~\citep{salemi2023lamp} leverages personalized information to boost performance on specific tasks such as review generation~\citep{li-tuzhilin-2019-towards}, recipe generation~\citep{majumder-etal-2019-generating}, and headline generation~\citep{ao-etal-2021-pens}.  This line of work requires model providers to make better models utilizing the personal information of the user. In our work, we propose a framework that allows users to choose which preference the language model should prefer, essentially giving control to the user. 

\paragraph{Parameter Merging}
Recent work has shown that performing weighted linear interpolation of model parameters leads to the composition of each model ability~\citep{li2022branch, wortsman2022model, wortsman2022fi, don2022cold, Huang2023LoraHubEC}. This line of work has led to many interesting applications of model merging such as composing the abilities of expert models that perform different tasks~\citep{ilharco2022editing, pmlr-v202-jang23a} and introducing language-specific modules for growing the total capacity of multilingual LMs~\citep{pfeiffer2022lifting}.

Most recently, \citet{rame2023rewarded} proposed to merge policy models that were trained to perform specific tasks such as question answering and summarization using proxy reward models. While they mostly deal with reward models trained on the same data, our proposed MORL methods are an extension of this work that actually deals with diverse reward models trained on multifaceted human feedback to show compositional abilities through parameter merging rather than just ensembling.

\section{Reinforcement Learning from \textit{Personalized} Human Feedback}
The current RLHF can be denoted as optimizing policy $\pi$:
\begin{equation}\label{eq1}
\pi^{*} = \operatorname*{arg\,max}_\pi(R(x,\pi(x)))
\end{equation}

where $R$ is the reward model trained on \textit{general} human feedback. As pointed out in \citet{silver2021reward}, the following optimization may implicitly be occurring under the hood:
\begin{equation}\label{eq2}
\pi^{*} = \operatorname*{arg\,max}_\pi(r_{1} + r_{2} + ... + r_{n})
\end{equation}

where $r_{i}$ represents rewards from objectives that human annotators may \textit{generally} consider positive objectives (e.g., informative, helpful, kind, etc.) and $n$ is the total number of unique `dimensions' of these positive objectives. This formulation does not allow modeling \textit{conflicting} objectives, which may occur in real-world scenarios. For example, some people may prefer concise and unpretentious responses in contrast to informative, polite responses.

In this section, we formalize RL$\mathcal{P}$HF where we allow modeling conflicting preferences during alignment. We explain how we collect \textit{conflicting} feedback in Section \ref{subsec:setup}. In Section \ref{subsec:morl}, we explain how we convert the current RLHF formulation into a MORL problem. Lastly, we explain the details of our evaluation in Section \ref{subsec:eval}.

\subsection{Personalized Reward and Feedback}
\label{subsec:setup}
\paragraph{Collecting \textit{Conflicting} Pairwise Feedback}
We utilize Tulu-7B LM~\citep{wang2023far}, a model that uses LLaMA-7B~\citep{touvron2023llama} as a base model and is instruction tuned on a mixture of open-source instruction-tuning datasets, as the base model for our experiments. We utilize 10k prompt instances from GPT4-Alpaca~\citep{peng2023instruction}, one of the datasets used to train Tulu-7B, as our instruction dataset $D_{train}$ to generate rollouts and collect pairwise feedback data. We also use the same $D_{train}$ during Proximal Policy Optimization (PPO) training~\citep{schulman2017proximal} of Tulu-7B. 

Following previous work, we simulate human annotators with GPT-4 for collecting large-scale pairwise feedback data~\citep{bai2022constitutional, dubois2023alpacafarm}---but note that our evaluations are validated with (smaller-scale) human preference data collected from crowdworkers. While \citet{dubois2023alpacafarm} mostly simulates GPT-4 and other LLMs to choose which is \textit{generally} a better response between two candidate responses, we provide GPT-4 with a single preference (full list shown in Table \ref{table:lisf_of_prefs}) to decide which is a better response. We also provide the same preference criteria via additional prompts during the rollout generation of the two candidate responses; we use Tulu-30B for the rollout generation while the actual policy model we train is Tulu-7B for our main experimental setup, making our experimental setting an off-policy training set-up.

\begin{table*}[t!]
\centering
\fontsize{7}{9}\selectfont
\begin{tabular}{llc}
    \toprule
    Dimension & Preference Prompt & Symbol\\
    \midrule
    \multirow{2}{*}{Expertise} & Generate/Choose a response that can be easily understood by an elementary school student. & P1A\\
    & Generate/Choose a response that only a PhD Student in that specific field could understand. & P1B\\ \midrule
    \multirow{2}{*}{Informativeness} & Generate/Choose a response that is concise and to the point, without being verbose. & P2A\\
    & Generate/Choose a response that is very informative, without missing any background information. & P2B\\ \midrule
    \multirow{2}{*}{Style} & Generate/Choose a response that is friendly, witty, funny, and humorous, like a close friend. & P3A\\
    & Generate/Choose a response (that answers) in an unfriendly manner. & P3B\\
    \bottomrule        
\end{tabular}
\caption{List of 6 conflicting preferences divided into 3 distinct dimensions that are used for our experiments. We simulate a scenario where a person has a preference from each dimension, resulting in 8 unique combinations of preferences: AAA, AAB, ABA, ABB, BAA, BAB, BBA, and BBB.}
\label{table:lisf_of_prefs}
\end{table*}  

\paragraph{Reward Model Training}
\begin{figure}[t] 
\includegraphics[width=\textwidth]{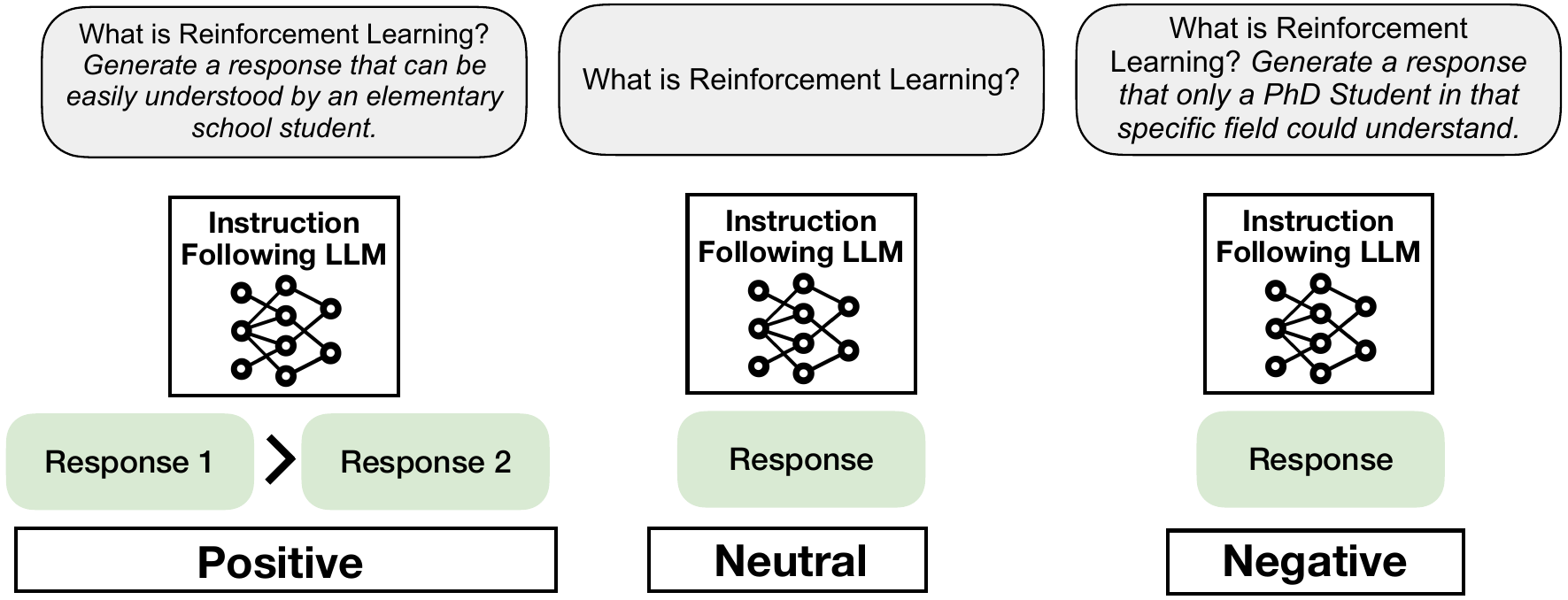}
\caption{Reward model training with conflicting preferences.}
\label{fig:rm_training}
\end{figure}
While we have feedback on which of the two model responses is more aligned with a single preference via GPT-4 annotation, utilizing only two \textit{positive} pairs during reward model training was empirically shown to be less robust during the PPO training. Instead, we train our reward model on multiple comparisons~\citep{song2023preference,kim2023aligning} by including a \textit{neutral} response and a \textit{negative} response as shown in Figure \ref{fig:rm_training}. Specifically, the reward model is provided with four different comparisons for a single prompt during training: positive 1 $>$ positive 2 (decided by GPT-4), positive $>$ neutral, positive $>$ negative, and neutral $>$ negative. The positive response when compared with the neutral and the negative response is chosen randomly. This allows the reward model to be exposed to different granularity of the specific preference and give scores accordingly during PPO training. We explore (1) training a single reward model in a multitask fashion that leverages the preference prompts during inference to give distinct rewards according to each preference and (2) training multiple reward models, each tailored to the distinct preference.

\subsection{Multi-Objective Reinforcement Learning (MORL)}
\label{subsec:morl}
The MORL problem can be denoted as:
\begin{equation}\label{eq3}
\pi^{*} = \operatorname*{arg\,max}_\pi(w_{1}r_{2} + w_{1}r_{2} + .. w_{n}r_{n})
\end{equation}

where $\pi^{*}$ is the policy model that aims to maximize multiple objectives from the rewards $r_{1}, r_{2}, ..., r_{n}$ and $w_{1}, w_{2}, ..., w_{n}$ is the importance placed on each objective. If $w_{1}, w_{2}, ..., w_{n}$ are constants during PPO training, this problem essentially becomes a single objective problem, maximizing towards a single, \textit{general} objective. In our setup, we have \textit{conflicting} preferences which require dynamically varying $w_{i}$ in a binary manner with respect to the conflicting preference during training and inference.

\paragraph{\textsc{Prompted-MORL (\textsc{P-Morl})}} First, we introduce a strong baseline that varies $w_{i}$ during MORL through \textit{prompts}. While $w_{i}$ are given as inputs directly to the policy model in traditional RL settings using PPO with MORL, there is no straightforward way of integrating different $w_{i}$ as an input to LLMs. Instead, we utilize the preference prompts as binary signals for $w_i$. 

We append the unique preference combination (shown in Table \ref{table:lisf_of_prefs}) with a training prompt $t$ from $D_{train}$ \{$t$ + \textit{P1} + \textit{P2} + \textit{P3}\}~\footnote{\textit{P1}, \textit{P2}, \textit{P3} each represents preference prompts from each preference dimension in Table \ref{table:lisf_of_prefs}. For one example, one unique combination might be P1A + P2B + P3A (ABA) where the combined objective for the response needs to be elementary level, informative, and friendly.} before feeding it to our initial policy model and getting the output response. Then, we gather reward signals for each of the preferences by feeding \{$t$ + \textit{P1/P2/P3} + output\} into a single reward model (doing three forward passes)~\footnote{Empirically, utilizing a single reward model instead of multiple reward models led to better performance. We hypothesize this is due to the problem of normalizing signals from different reward models~\citep{hayes2022practical}, which is known to be a nontrivial problem in MORL.} to get the reward signal specific to the individual preference and averaging the three reward values to get a single scalar reward used for PPO training. We multitask train the policy model across the eight different unique combinations of preferences, which essentially results in varying $w_{i}$. 

While \textsc{Prompted-MORL} can be a clear baseline for converting the alignment problem into a MORL problem, we propose another approach that does not have to see all existing preference combinations during training thus allowing increasing the total number of distinct preferences at scale, which is required for true personalization.

\paragraph{\textsc{Personalized Soups (P-Soups)}}
We decompose the MORL problem into multiple single-objective problems:
\begin{equation}\label{eq4}
\pi_{1}^{*} = \operatorname*{arg\,max}_\pi(r_{1}), \pi_{2}^{*} = \operatorname*{arg\,max}_\pi(r_{2}), ... , \pi_{n}^{*} = \operatorname*{arg\,max}_\pi(r_{n})
\end{equation}

where we optimize each policy $\pi_{n}$ individually. Then during inference, we pick and choose the policy models whose objective we want to maximize together and perform a weighted sum of the parameters on the fly:

\begin{equation}\label{eq5}
\pi^{*} = w_{1}\pi_{1}^{*} + w_{2}\pi_{2}^{*} + ... + w_{n}\pi_{n}^{*}
\end{equation}

where $\sum_{1}^{n}w_{n} = 1$.

This means that even though the exact preference combinations haven't been seen during training, we are still able to composite them on the fly during inference. While the total computational complexity increases exponentially when we are required to observe all possible combinations, optimizing individual objectives separately only increases the complexity in linear space.

This also means that multitask training is not necessary and allows efficient integration of novel preferences. Since \textit{personalization} also entails that there can be an infinite number of new preference dimensions, we assert that \textsc{Personalized Soups} makes tackling RL$\mathcal{P}$HF feasible.

\subsection{Multifaceted Evaluation}
\label{subsec:eval}
\paragraph{Evaluation}
For evaluation, we manually filter out 50 instances from the Koala evaluation~\citep{geng2023koala} that require open-ended generations. We also modified some of the prompts so that the evaluation prompts do not contain any elements requiring individual preferences (e.g., removing the phrase asking for a elementary-level response from the original prompt since we want to test the LLM to generate a expert-level response). The full list of evaluation prompts used for our experiments is shown in Appendix \ref{appen:eval_prompts}. In our evaluation setup, we simulate users to have a unique combination of preferences, each from the three preference dimensions (Expertise, Informativeness, Style) in Table \ref{table:lisf_of_prefs}, which equates to 8 unique preference combinations (examples shown in Figure \ref{fig:main}). We get the average win rate across the simulated 8 preference combinations for our final evaluation. We use a variant of the AlpacaFarm evaluation framework for simulated (GPT4) evaluation and hire 24 crowdworkers for human evaluation. Details of human evaluation are provided in Appendix \ref{appen:eval}.

\paragraph{Aggregated Win Rate}
When given an evaluation prompt $p$, we first generate responses from each model to get the outputs $\theta_{A}(p) = o_{A}, \theta_{B}(p) = o_{B}$ where $\theta_{A}$ is model A and $\theta_{B}$ is model B. The common approach is get $H(p, o_{A}, o_{B}) = $ \{\textsc{Win}, \textsc{Tie}, \textsc{Lose}\} by asking the human which model response is \textit{generally} preferred.

In our evaluation setup, we first assign scores to each of the possible feedback: \textsc{Win} = 1, \textsc{Tie} = 0, \textsc{Lose} = -1. Next, we iterate through the different preference dimensions and get an aggregated score value: $\sum_{i=1}^{n}H_{i}(p, o_{A}, o_{B}) = score$.
%where $w_{i}$ represents the importance to place on the $i$th preference dimension, which is usually set to 1 as default, if not stated otherwise.
Finally, we have \textsc{Win} if $score>0$, \textsc{Tie} if $score=0$, and \textsc{Lose} if $score<0$. To get the final win rate between $\theta_{A}$ vs. $\theta_{B}$, we iterate through the entire evaluation set (50 prompts) $\times$ the unique preference combinations (8 combinations) and get the total $\frac{\#Wins}{\#Wins + \#Loss}$ as the final win rate, while disregarding the total number of ties.

\section{Experiments}
\subsection{Baseline Methods}
\label{subsec:baseline}
\begin{table*}[ht!]
\centering
\fontsize{9}{11}\selectfont
\begin{tabular}{lccc}
\toprule
Method & Reward Model & Policy Model & Training \\
\midrule
Vanilla Baseline & - & single & - \\
Traditional RLHF & single & single & single \\
Preference Prompting & - & single & - \\
Multi-task Training & single & single & multitask \\
\midrule
Prompted MORL & single/multi & single & multitask  \\
\textsc{Personalized Soups} & multi & multi & single  \\
\bottomrule        
\end{tabular}
\caption{Components of different methods.} 
\label{table:method_components}
\end{table*}  
In this subsection, we provide details of the single-objective baseline methods we implement. The summary of the key component differences in comparison with our proposed methods is provided in Table \ref{table:method_components}.

\textbf{Vanilla Baseline (VB)} As the most simple baseline, we simply utilize the base Tulu-7B model to generate responses without providing it any notion of preferences. During the evaluation, we use the same response to evaluate on the 8 different preference combinations. 

\textbf{Reinforcement Learning from Human Feedback (RLHF)} We perform RLHF in the traditional manner where GPT-4 labels which response is \textit{generally} better, train a reward model using the pairwise feedback data, and use the reward model to adapt the policy model with PPO training. The same 10k instances from $D_{train}$ are used for RLHF. 

\textbf{Preference Prompting (PP)} Next, we observe how far the instruction-tuned base LM can integrate multiple preference combinations by simply prompting for the preferences without any additional training. 

\textbf{Multi-task Training (MT)} For a competitive baseline, we utilize the positive candidate selected by GPT-4 as the output for imitation learning, which is essentially performing rejection sampling~\citep{Nakano2021WebGPTBQ} that uses GPT-4 as the reward model for selecting golden responses from the distribution of responses. We append the individual preference prompt with instances from $D_{train}$ and multitask train the Tulu-7B model across all six individual preferences. This method also allows \textit{distilling} the outputs of Tulu-30B for training Tulu-7B.

\subsection{Experimental Details}
For both the reward model and policy training, we limit ourselves to going through $D_{train}$ only once (1 epoch). In the initial exploration stage, the end performance for the policy model did not improve even if we trained the reward model for longer epochs. For policy model training, we utilize our evaluation dataset (50 prompts) to get the average reward and chose the policy model checkpoint that showed the highest average reward on the evaluation set for our final evaluation. We utilize LoRA~\citep{hu2022lora} for both the reward model and policy model training. The detailed hyperparameters for the reward model and policy model training are provided in our github repository~\footnote{\url{https://github.com/joeljang/RLPHF}}.

\subsection{Main Results}
\begin{table*}[t!]
\fontsize{10}{12}\selectfont
\centering
\begin{tabular}{l|cccccc|c}
\toprule
Method & VB & RLHF & PP & MT & \textsc{P-Morl} & \textsc{P-Soups} & Avg.\\
\midrule
VB & - & 44.52 & 45.95 & 43.00 & 46.58 & 40.14 & 44.04\\
RLHF & 55.48 & - & 37.81 & 38.98 & 40.48 & 45.09 & 43.57\\
PP & 54.05 & 62.19 & - & 48.08 & 49.09 & 45.00 & 51.68\\
MT & 57.00 & 61.02 & 51.92 & - & 48.37 & 46.64 & \underline{52.99} \\
\midrule
\textsc{P-Morl} & 53.42 & 59.52 & 50.91 & 51.63 & - & 46.96 & 52.49 \\
\textsc{P-Soups} & 59.86 & 54.91 & 55.00 & 53.36 & 53.04 & - & \textbf{55.23} \\
\bottomrule        
\end{tabular}
\caption{Simulated pairwise win rate (\%) across all methods using GPT-4.} 
\label{table:gpt4_eval}
\end{table*} 

\begin{table*}[t!]
\fontsize{10}{12}\selectfont
\centering
\begin{tabular}{l|cccccc|c}
\toprule
Method & VB & RLHF & PP & MT & \textsc{P-Morl} & \textsc{P-Soups} & Avg.\\
\midrule
VB & - & 46.56 & 47.16 & 41.40 & 37.29 & 39.39 & 42.36\\
RLHF & 53.44 & - & 52.73 & 44.22 & 38.18 & 44.97 & 46.71 \\
PP & 52.84 &  47.27 & - & 42.33 & 43.75 & 37.50 & 44.74\\
MT & 58.60 & 55.78 & 57.67 & - & 43.48 & 45.45 & 52.20\\
\midrule
\textsc{P-Morl} & 62.71 & 61.82 & 56.25 & 56.52 & - & 56.21 & \textbf{58.70}\\
\textsc{P-Soups} & 60.61 & 55.03 & 62.50 & 54.55 & 43.79 & - & \underline{55.29}\\
\bottomrule        
\end{tabular}
\caption{Pairwise win rate (\%) across all methods through Human Evaluation.} 
\label{table:human_eval}
\end{table*} 

Table \ref{table:gpt4_eval} and \ref{table:human_eval} show the results of doing all possible pairwise comparisons across the methods using GPT-4 and humans as judges, respectively. Note that the win rate of each battle is calculated using the aggregated win rate explained in Section \ref{subsec:eval}. Each individual preference combination results are shown in Appendix \ref{appen:full_results}. We also show the average \textit{criteria-wise} win rate instead of the \textbf{aggregated} win rate across all of the methods in Appendix \ref{appen:criteriawise}.

The first thing to note is that there is a limitation to the extent prompting (PP) can integrate multiple preferences. This means that specific training for integrating the multiple preferences is necessary to composite them together. Next, we can see that supervised fine-tuning (RS) underperforms MORL-based methods, which is consistent with prior work that also showed the advantage of RL-based approaches when aligning LLMs with human feedback compared to its supervised-finetuning counterpart. Finally, while \textsc{P-Morl} and \textsc{P-Soups} both outperform other methods on average, there exists a discrepancy between the simulated and human evaluation; \textsc{P-Soups} has the highest average win rate in GPT-4 evaluation while \textsc{P-Morl} has the highest in human evaluation. Nonetheless, \textsc{P-Soups} is able to show superior performance in comparison to baseline methods and competitive performance to \textsc{P-Morl}.

In previous parameter merging literature, multi-task fine-tuning (RS) used to be considered the upper bound for compositional abilities through parameter merging~\citep{don2022cold}. However, in our scenario, we can see that parameter merging (\textsc{P-Soups}) is able to \textit{outperform} multitask fine-tuning, showing promising results for parameter merging not only as a distributed multitask finetuning method but a method that can result in superior performance than multitask training. 

\paragraph{Trade-off Between General \textit{Helpfulness}}
\begin{figure}[t!] 
\centering
\includegraphics[width=0.9\textwidth]{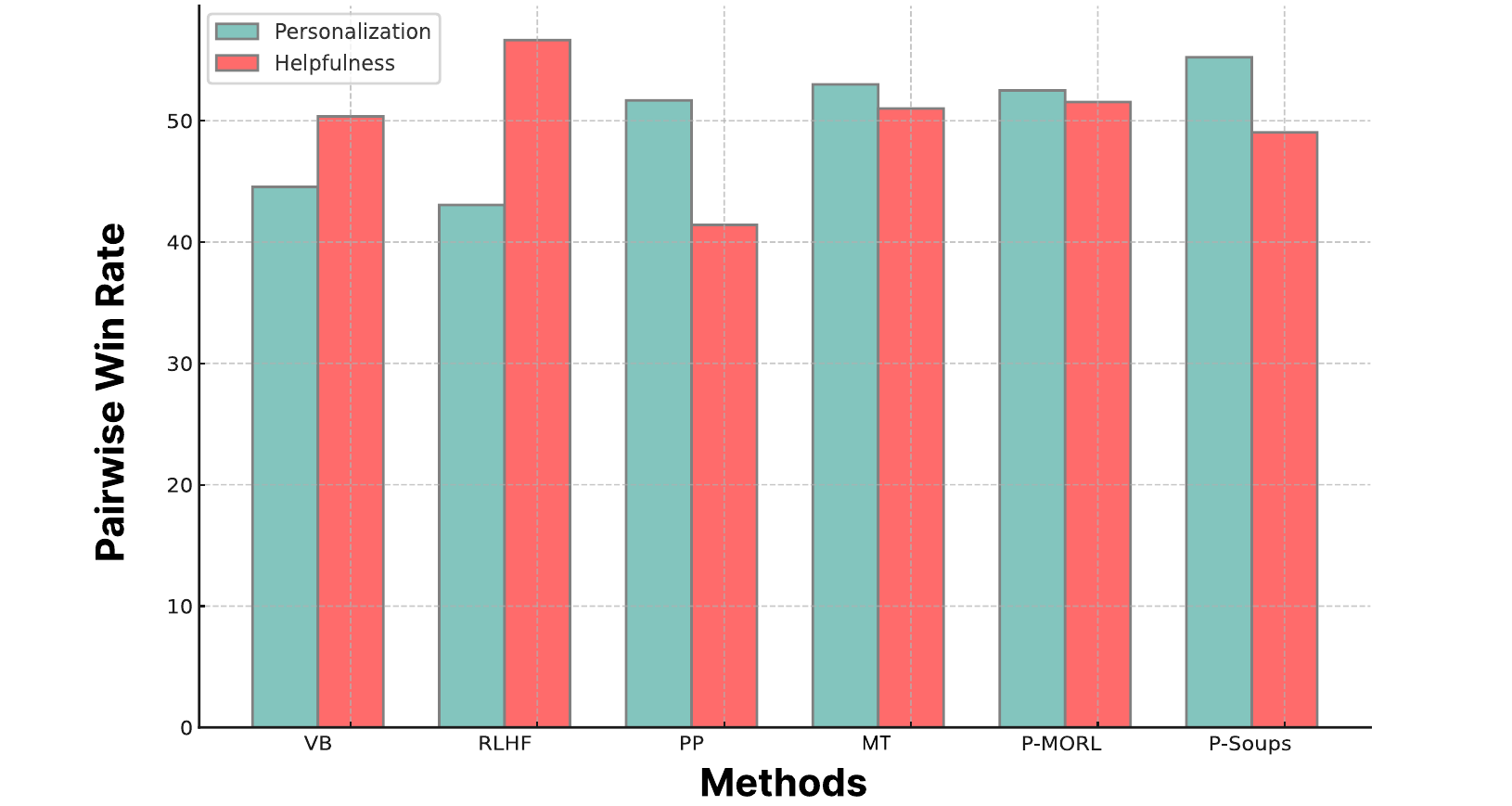}
\caption{Average simulated pairwise win rate (\%) across all the methods using personalization and helpfulness as the evaluation criteria.}
\label{fig:helpfulness}
\end{figure}
One might still wonder about the general `helpfulness' capabilities of models that are trained to be tailored to multiple preferences. In Figure \ref{fig:helpfulness}, we first show the average pairwise win rate from Table \ref{table:gpt4_eval} in green. Next, we instead ONLY perform pairwise comparisons with the unseen objective `helpfulness' (ask GPT-4 which model response they \textit{generally} prefer better) and report the average pairwise win rate in red. 

RLHF performs the best in this scenario, which shows that there is \textit{no free lunch}; while the objective of RLHF was to provide model responses that are \textit{generally} preferred (highly correlated with `helpfulness'), the other methods were prompted/trained to be optimized towards the personalized preference aspects, possibly deviating away from general helpfulness. While RS, \textsc{P-Morl}, and \textsc{P-Soups} are able to retain similar performance in terms of helpfulness compared to the initial instruction-tuned model (VB), we observe that prompting (PP) significantly underperforms compared to other methods which also highlights the limitation of simply prompting base/instruction-tuned models for personalized preferences and shows the need for specialized training methods for personalization. 

\subsection{Scaling to New Preferences}
\begin{figure}[t!] 
\centering
\includegraphics[width=1.0\textwidth]{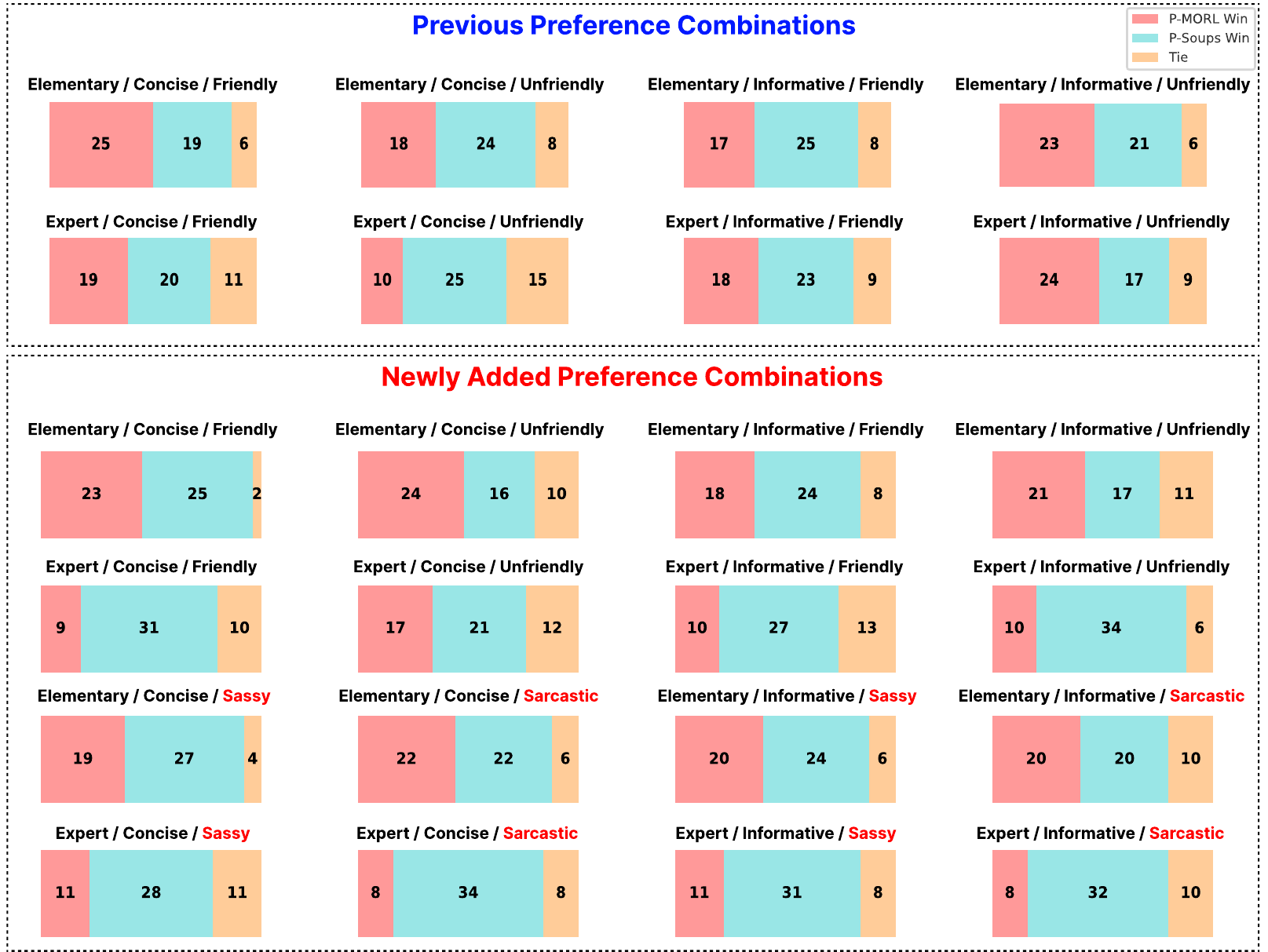}
\caption{The \textit{win / lose / tie} statistics for each of the \textit{original} and \textit{new} preference combination using GPT-4 as the evaluator. Getting the total $\frac{\#Wins}{\#Wins + \#Loss}$ for the 16 preference combinations results in \textbf{59.06\%} win rate for \textsc{P-Soups} over \textsc{P-Morl}.}
\label{fig:additional}
\end{figure}
While we explore 6 distinct preferences in this work, we are still limited in doing `declarative' personalization; that is, the individual preferences have been pre-defined to measure the performance of different methodologies. However, in the real world, individuals may not be bound by pre-defined preferences. Furthermore, people's preferences might change over time, which requires continual learning of new preferences. This means that we may be required to train \textit{infinite} numbers of preferences to be truly personalized to individuals' preferences. Considering this aspect, the \textit{scalability} of methods becomes a critical factor in implementing RLPHF in real-world scenarios. 

In order to compare the scalability of \textsc{P-Morl} and \textsc{P-Soups}, we add two new preferences (in addition to the ones in Table \ref{table:lisf_of_prefs} to the \textsc{Style} dimensions: (P3C) ``\textit{Generate/Choose a response (that answers) in a sassy manner.}" and (P3D) ``\textit{Generate/Choose a response (that answers) in a sarcastic manner.}", which results in a total of 16 (2 $\times$ 2 $\times$ 4) unique preference combinations. We re-train \textsc{P-Morl} on the 16 new preference combinations and only train two new policy models for integrating \textsc{P-Soups}. The simulated win-rate between \textsc{P-Morl} and \textsc{P-soups} on each of the original preference combinations (53.04\% win rate of \textsc{P-Soup} over \textsc{P-Morl} in Table \ref{table:gpt4_eval} decomposed into each preference combinations) and the 16 new preference combinations are shown in Figure \ref{fig:additional}.

As shown in the figure, \textsc{P-Soups} shows competitive performance compared to \textsc{P-Morl} while being much more efficient considering that it (1) did not have to observe all 16 possible preference combinations and (2) did not have to re-train on the previous preferences, but just train two new policies each for the new preference in a modular manner and merge their parameters on-the-fly during inference. Considering that \textsc{P-Morl} is bounded by $O(2^n)$ while \textsc{P-Soups} is bounded by $O(n)$ where $n$ is the total number of preferences (assuming there are two unique preferences for each dimension), we assert that \textsc{P-Soups} allows tackling RLPHF to be feasible. 

%When simulating a scenario to see the relative computational complexity between \textsc{P-Morl} and \textsc{P-Soups} when the total number of preference dimensions increases (shown in Figure \ref{fig:scaling}), assuming there are only two atomic preferences in each dimension, \textsc{P-Morl} increases by $O(2^n)$ while \textsc{P-Soups} increases by $O(n)$. This highlights the scalability of \textsc{P-Soups} for real-world scenarios where users may have an infinite number of preference combinations. We show empirical output generations of \textsc{P-Soups} of each of the 16 preference combinations in Appendix \ref{appen:examples}.

%It is important to maintain the results of the \justblue{Previous Preference Combinations} while \textit{efficiently} adapting on \justred{Newly Added Preference Combinations}. \textsc{P-Soups} satisfies both conditions across different combination settings in a more efficient manner compared to \textsc{P-MoRL} (\textsc{P-Soups}: $O(n)$ \textit{vs} \textsc{P-MoRL}: $O(2^n)$). The simulated pairwise win rate (as reported in Table~\ref{table:gpt4_eval}) is improved from 53.04 $\rightarrow$ 59.64.

%\begin{figure}[t!] 
%\centering
%\includegraphics[width=0.9\textwidth]{figures/complexity.pdf}
%\caption{Relative FLOP comparison between \textsc{P-Morl} and \textsc{P-Soups} in a simulated scenario where the total \# of preference dimension increases and there are two conflicting preferences in each dimension.}
%\label{fig:scaling}
%\end{figure}

\section{Conclusion}
Previous work has shown that adapting LLMs with RLHF helps them generate outputs that are preferred by humans over the supervised fine-tuned counterpart. However, recent work has also pointed out that simply training LLMs to abide by the preference of the general may result in ignoring individual preferences and values. In this work, we provide the first steps to tackle this issue by proposing Reinforcement Learning from \textit{Personalized} Human Feedback as a multi-objective problem so that LLMs can be aligned to follow \textit{conflicting} preferences. We propose a promising method called \textsc{P-Soups} that is able to composite models trained on single objectives on the fly during inference. We also highlight the scalability of \textsc{P-Soups} by showing that it scales linearly, instead of exponentially like the MORL baseline, with regards to the number of new preferences, which is required to provide true personalization to individual users.

\subsubsection*{Acknowledgments}
Thanks to Minyoung Hwang, Sungdong Kim, Tim Dettmers, Yoonjoo Lee, and Margaret Li for helpful feedback and discussion.

\bibliography{iclr2024_conference}
\bibliographystyle{iclr2024_conference}

\appendix
\section{Details of Evaluation Setup}
\label{appen:eval}
\paragraph{Simulated pairwise evaluation}
We use a modified version of the GPT4 annotation prompt used by \citet{dubois2023alpacafarm}. We modify the criteria to perform the pairwise evaluation from \textit{general} to a single preference dimension. We also provide 3 demonstrations: one scenario where there is a tie because both responses do not contain any notion of the preference (e.g. both responses do not show any signs of friendliness), one scenario where there is a clear winner, and one scenario where they are both good, but one is better than the other. 

\begin{figure}[h!] 
\centering
\begin{subfigure}[b]{\textwidth}
\includegraphics[width=\textwidth]{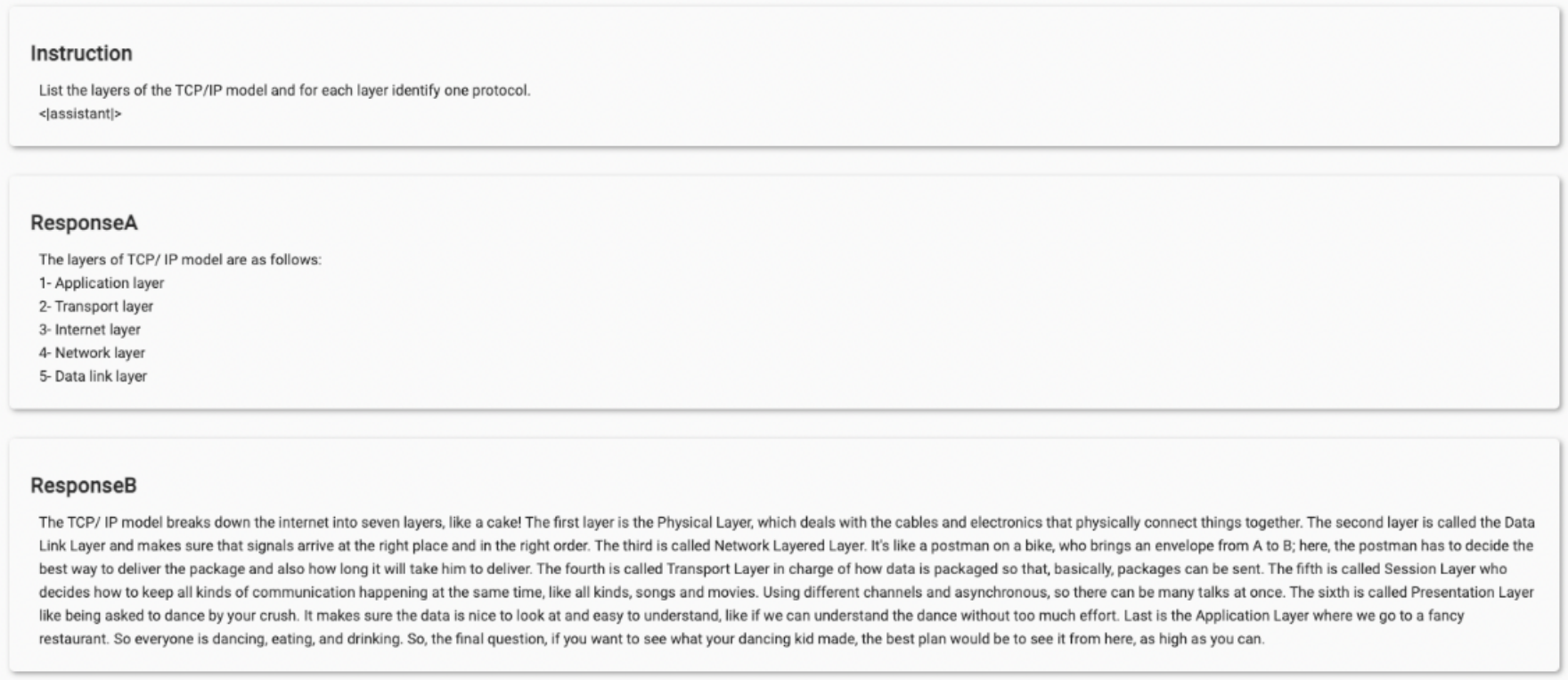}
\end{subfigure}
\begin{subfigure}[b]{\textwidth}
\includegraphics[width=\textwidth]{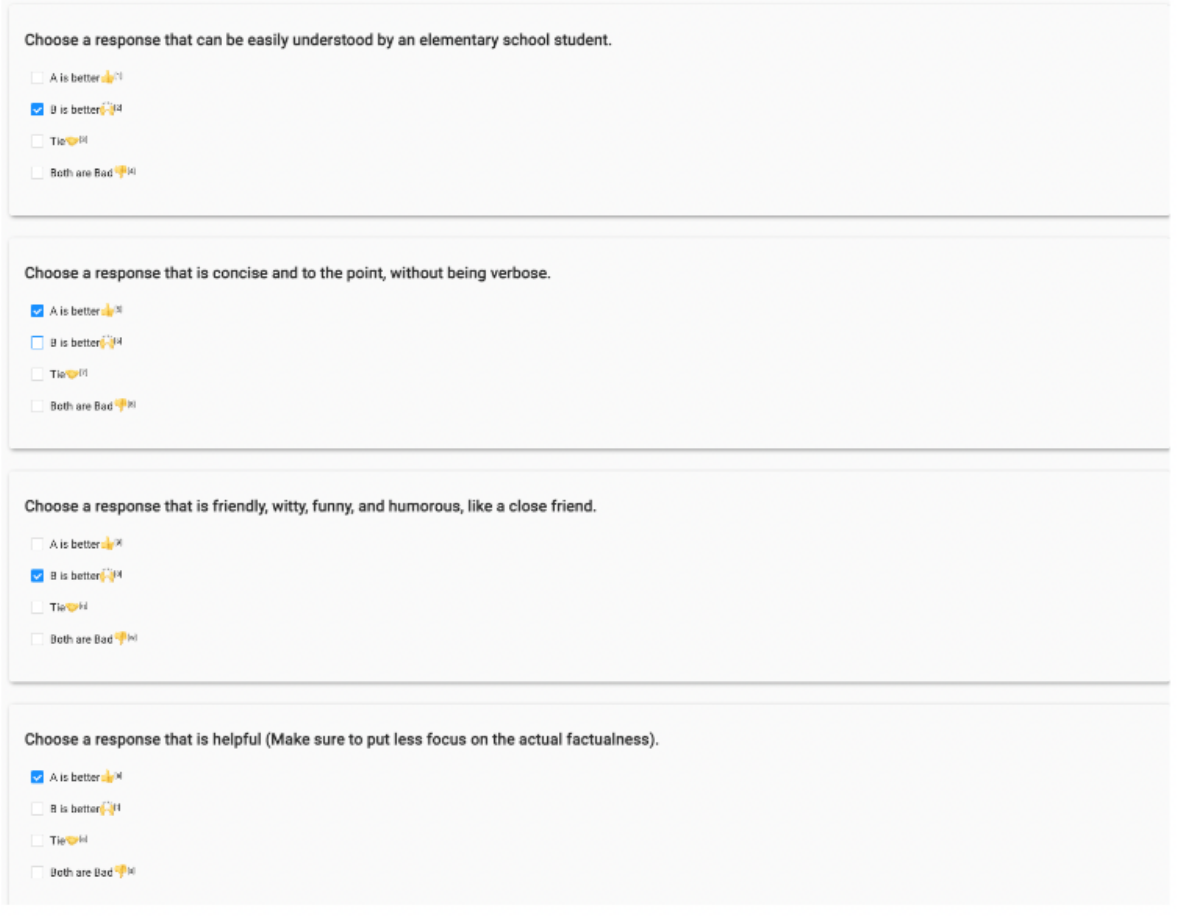}
\end{subfigure}
\caption{Interface for human evaluation}
\label{fig:human_interface}
\end{figure}

We recruited 24 crowd workers for our human evaluation. Figure \ref{fig:human_interface} shows the interface used for human evaluation. We consider both the `Tie' and `Both are bad' options to be Ties.

\section{Criteria-wise Evaluation}
\label{appen:criteriawise}
\begin{figure}[h!] 
\centering
\includegraphics[width=1.0\textwidth]{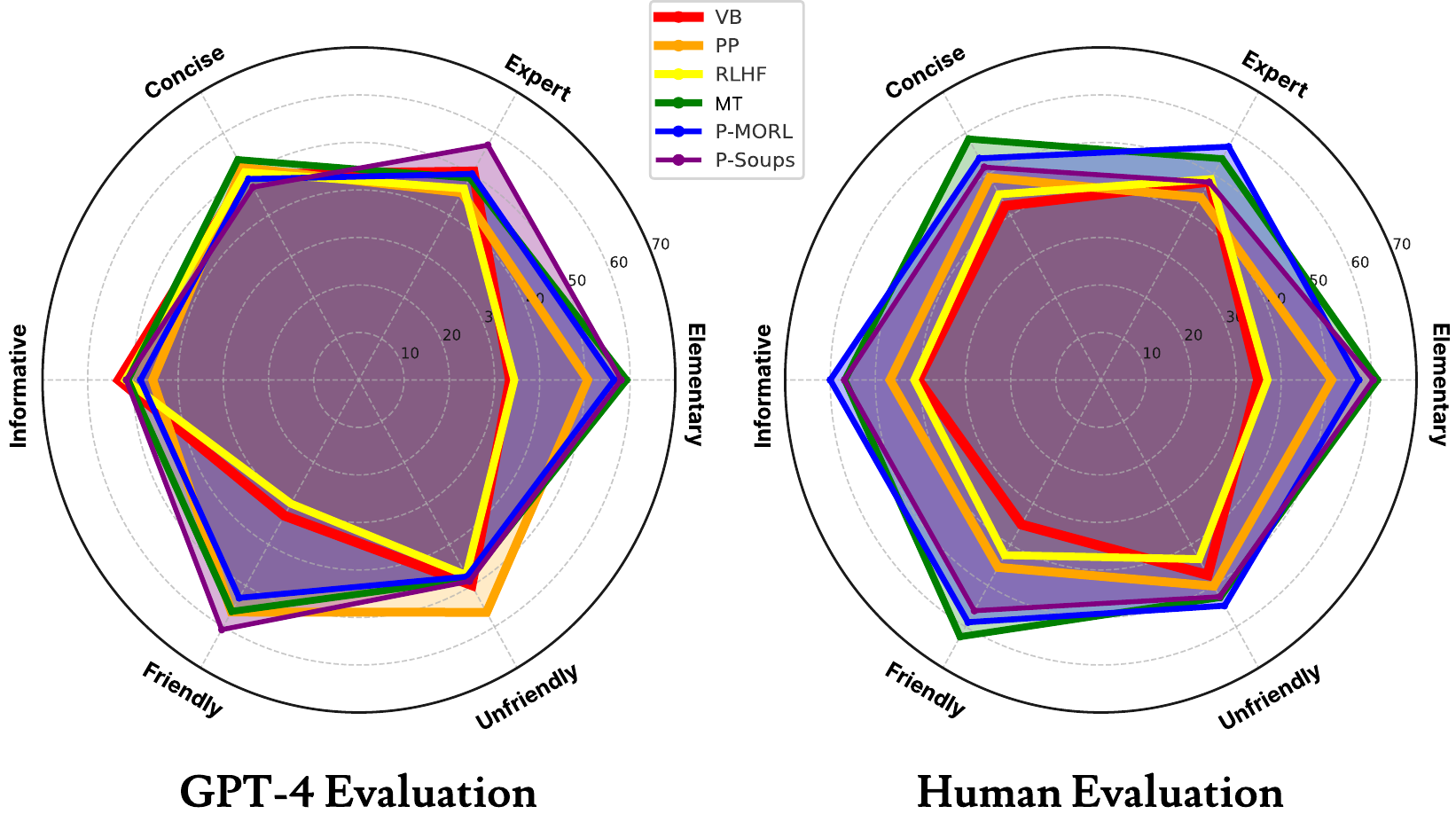}
\caption{The criteria-wise win rate (\%) across all methods through Human Evaluation and GPT-4 Evaluation.}
\label{fig:criteriawise_eval}
\end{figure}
The criteria-wise win rate (\%) across all the methods are shown in Figure \ref{fig:criteriawise_eval}. The criteria-wise win rate was calculated by getting the average win-rate of the preference combinations that contained the specific preference dimension. For example, when calculating the criteria-wise win rate of `Elementary', we got the average win-rate of the preference combinations that contained the `Elementary' preference, which includes AAA, AAB, ABA, and ABB. 

\section{The full list of evaluation prompts}
\label{appen:eval_prompts}
\begin{table}[h!]
\fontsize{9}{11}\selectfont
\centering
\begin{tabular}{cl}
\toprule
ID & Prompt \\ \midrule
1 & Why can't bank use cash as capital as a buffer for potential losses? \\
2 & Can you tell my a story about nuclear physics? \\
3 & Create a message for an offer letter for an internship at Apple as a AI research intern. \\
4 & Explain sarcoidosis to me. \\
5 & "Give me a sample 5 day itienary for a switzerland holiday, starting from Basel" \\
6 & Explain The Death of Ivan Ilych \\
7 & "Why is it that only proteins, carbohydrates and fats are deemed to have caloric value?" \\
8 & "Rank the following companies by how pro-consumer they are:Microsoft, Google, Nintendo, Sony, EA." \\
9 & I’m planning to visit Okinawa Japan from April 7th to April 10th. Do you have any recommendation on \\
& what to do while I’m there? \\
10 & How to improve instance segmentation AP metrics with human knowledge? \\
11 & What is needed for self-sufficient living spaces? \\
12 & Expand on the relation between inconsistency and ideological thinking. \\
13 & Why do people give Reddit Gold to admins? \\
14 & What does Nassim Nicholas Taleb say about market efficiency? \\
15 & Can a boy and girl be just be best friends only ever? \\
16 & What would be the reason for the popularity of youtube compared to other video sharing websites? \\
17 & "Do you know something about the book  "the art of thinking clearly" wrote by Rolf Dobelli? \\
18 & Antennas that transmit an equal amount of energy in the horizontal direction are called \\
19 & "Hi, I have a question about MFCC (mel frequency cepstral coefficients). Are they the same thing as a \\
& MEL-spectogram, or is there a difference?" \\
20 & Why is it a bad idea to give a mouse a cookie? \\
21 & How can anti-deressants make people think or act suicidally? \\
22 & Create a lesson plan in two different levels: in CEFR A1 and A2 on the topic of friendship. Add a rubric. \\
23 & Is online casino legal in India? \\
24 & "How much of a threat is climate change in the coming years, and what should we do to stop it?" \\
25 & Explain the basics of area and perimeter \\
26 & What are the possible performance issues in a learning program on cross cultural communication? \\
27 & Write description for 925 sterling silver miami cuban link chain. \\
28 & What if people only valued and took pride in the things that were true about themselves? \\
29 & I need to learn English could you help me and make a exercise? \\
30 & Why does warmth make you feel tired? \\
31 & Explain to me the Finite Elemente Method. \\
32 & Introduce the ”financial markets and institutions“ by Frederic S. Mishkin \\
33 & When are hops added to the brewing process? \\
34 & Can a Liebherr LTM 11200-9.1 hypothetically lift Mount everest? \\
35 & What are five important topics for game design? \\
36 & What language does argentina people speak? \\
37 & Is queue an ADT or a data structure? \\
38 & What are some basics of nutrition that i should be aware of? \\
39 & Can a qualifying manager work remotely abroad for US employer for an year be eligible for Eb1-C\\
& while on h1-b? \\
40 & "I have competencies in remote sensing, machine learning, and water resource knowledge, what are \\
& the possible jobs I can occupy? What are the possible projects I can do? What companies I can work at?" \\
41 & Is a banana a fruit or a herb \\
42 & What are african governments doing to improve air traffic connectivity? \\
43 & I want to open the developler tools in chrome with ctrl + shift + i on this website: https://mnsw.pro/ It doesnt work.\\
& works on other websites. even here. what is wrong? \\
44 & "Consider this situation. Customer is ready to pay \$13 for my product but we are not able to raise invoice for less \\
& than \$20. So in this situation, act like a top sales person; closing expert; give me an email copy to upsell this \\
& customer with another \$13 product so that it can create a win win situation." \\
45 & What are the important points for brand promotion and brand marketing? \\
46 & What niches are not being fulfilled on the YouTube platform? \\
47 & Explain TypeScript and Duck Typing. \\
49 & How are carbon fibers used in buildings? \\
50 & List the layers of the TCP/IP model and for each layer identify one protocol. \\
\bottomrule
\end{tabular}
\caption{The full list of 50 prompts used for evaluation.}
\label{table:DE_list}
\end{table}
The full list of evaluation prompts used in our experiments are provided in Table \ref{table:DE_list}.

\section{Detailed results for Human Evaluation and GPT-4 Evaluation}
\label{appen:full_results}
We provide detailed results (\textit{win / loss / tie} for each of the preference combinations for our main experimental results. Table \ref{table:gpt4_eval_detailed} shows the GPT4 evaluation and Table \ref{table:human_eval_detailed2} shows the human evaluation results.

\begin{table*}[t!]
\fontsize{6}{8}\selectfont
\centering
\begin{tabular}{l|cccccccc|cc}
\toprule
Method & AAA & AAB & ABA & ABB & BAA & BAB & BBA & BBB & Total. & Avg Win-rate\\
\midrule
VB \textit{vs} RLHF &15/29/6 & 17/27/6 & 12/21/17 & 15/17/18 & 18/20/12 & 21/22/7 & 20/14/16 & 16/17/17 & 134/167/99 & 44.52\%\\
VB \textit{vs} PP & 17/30/3 & 17/30/3 & 19/16/15 & 11/13/26&22/17/11 & 19/23/8 & 19/16/15 & 18/22/10 & 142/167/91 & 45.95\%\\
VB \textit{vs} MT & 11/35/4 & 15/30/5 & 14/20/16 & 17/11/22 & 25/19/7 & 24/21/5 & 9/16/25 & 14/19/17 & 129/171/100 & 43.00\%\\
VB \textit{vs} \textsc{P-MORL} & 20/30/0 & 19/30/11 & 14/18/12 & 12/17/21 & 18/23/9 & 25/20/5 & 20/11/19 & 15/15/20 & 143/164/93 & 46.73\%\\
VB \textit{vs} \textsc{P-Soups} & 18/30/2 & 18/27/5 & 9/23/18 & 7/20/23 & 16/24/10 & 19/24/7 & 13/12/25 & 16/13/21 & 116/173/111 & 40.14\%\\
RLHF \textit{vs} PP &11/38/1 & 15/32/3 & 9/24/15 & 15/20/15 & 17/23/10 & 23/25/2 & 18/15/17 & 13/22/15 & 121/199/80 & 37.81\%\\
RLHF \textit{vs} MT & 10/38/2 & 16/28/6 & 6/22/12 & 14/11/25 & 22/16/12 & 20/27/3 & 12/17/11 & 15/21/14 & 115/180/105 & 38.98\%\\
RLHF \textit{vs} \textsc{P-MORL} & 14/27/9 & 14/30/6 & 10/21/19 & 10/15/25 & 17/21/12 & 24/20/6 & 17/18/15 & 11/19/20 & 117/171/102 & 40.48\%\\
RLHF \textit{vs} \textsc{P-Soups} & 16/28/6 & 16/29/5 & 10/21/19 & 12/22/16 & 25/15/10 & 23/24/3 & 17/13/20 & 19/16/15 & 138/168/94 & 45.09\%\\
PP \textit{vs} MT & 19/28/3 & 26/18/6 & 16/21/13 & 12/22/16 & 23/20/7 & 25/22/3 & 12/17/21 & 18/15/17 & 151/163/86 & 48.08\%\\
PP \textit{vs} \textsc{P-MORL} & 23/23/6 & 31/15/4 & 17/17/16 & 13/27/10 & 22/22/6 & 26/17/7 & 14/24/12 & 16/23/11 & 162/168/70 & 49.09\%\\
PP \textit{vs} \textsc{P-Soups} & 20/27/3 & 26/21/3 & 15/20/15 & 12/25/13 & 15/24/11 & 23/21/6 & 16/21/13 & 17/17/16 & 144/176/80 & 45.00\%\\
MT \textit{vs} \textsc{P-MORL} & 24/21/5 & 24/25/1 & 19/18/13 & 16/14/20 & 20/17/13 & 19/26/5 & 16/14/20 & 19/12/19 & 157/147/96 & 48.37\%\\
MT \textit{vs} \textsc{P-Soups} & 25/22/3 & 13/23/6 & 21/27/12 & 22/22/6 & 23/18/9 & 17/19/14 & 18/20/12 & 13/23/14 & 152/174/74 & 46.64\%\\
\textsc{P-MORL} \textit{vs} \textsc{P-Soups} & 25/19/6 & 18/24/8 & 17/25/8 & 23/21/6 & 19/20/11 & 10/25/15 & 18/23/9 & 24/17/9 & 154/174/72 & 46.96\%\\
\bottomrule        
\end{tabular}
\caption{Simulated pairwise win rate (\%) across all methods using GPT-4.} 
\label{table:gpt4_eval_detailed}
\end{table*} 

\begin{table*}[t!]
\fontsize{6}{8}\selectfont
\centering
\begin{tabular}{l|cccccccc|cc}
\toprule
Method & AAA & AAB & ABA & ABB & BAA & BAB & BBA & BBB & Total.& Avg Win-rate\\
\midrule
VB \textit{vs} RLHF &9/18/3&12/13/5&10/15/5&11/13/6&13/12/5&13/4/13&6/14/10&14/12/4&88/101/51&46.56\%\\
VB \textit{vs} PP &13/11/6 & 5/17/8 & 9/12/9 & 16/11/3 & 3/12/15 & 14/11/5 & 9/11/10 & 14/8/8 & 83/93/64 & 47.16\%\\
VB \textit{vs} MT &4/22/4 & 5/20/5 & 4/18/8 & 10/6/14 & 7/13/10 & 12/9/9 & 10/17/3 & 6/13/11&58/118/64 &32.95\%\\
VB \textit{vs} \textsc{P-MORL} &6/21/3 & 12/15/3 & 7/19/4 & 5/15/10 & 11/10/9 & 15/9/6 & 11/13/6 & 6/8/16&73/110/57 & 39.89\%\\
VB \textit{vs} \textsc{P-Soups} &9/14/7 & 6/14/10 & 2/20/8 & 10/14/6 & 10/8/12 & 11/6/13 & 7/15/8 & 12/14/4&67/105/68 & 38.95\%\\
RLHF \textit{vs} PP &12/15/3&19/7/4&22/7/1&12/7/11&13/4/13&14/6/10&10/6/14&12/14/4&82/98/60 & 45.56\%\\
RLHF \textit{vs} MT &5/24/1&12/15/3&3/17/10&6/14/10&10/14/6&13/10/7&8/16/6&6/14/10&63/124/53 & 33.69\%\\
RLHF \textit{vs} \textsc{P-MORL} &3/16/11&8/10/12&8/13/9&6/10/14&7/9/14&13/13/4&5/14/11&10/17/3&60/102/78 & 37.04\%\\
RLHF \textit{vs} \textsc{P-Soups} &13/11/6&4/21/5&3/26/1&10/8/12&12/6/12&11/8/11&15/9/6&10/13/7&78/102/60 & 43.33\%\\
PP \textit{vs} MT & 7/16/7&8/11/11&7/12/11&9/10/11&7/6/17&5/9/16&6/14/10&11/9/10&60/87/93 & 40.82\%\\
PP \textit{vs} \textsc{P-MORL} &8/17/5&18/5/7&10/11/9&4/10/16&12/12/8&6/12/12&12/15/3&7/17/6&77/99/64 & 43.75\%\\
PP \textit{vs} \textsc{P-Soups} &12/14/4&16/12/2&8/20/2&10/11/9&6/11/13&8/9/13&7/9/14&8/6/16&75/92/73 & 44.91\%\\
MT \textit{vs} \textsc{P-MORL} &14/13/3&16/10/4&13/16/1&12/9/9&10/4/16&6/14/10&10/15/5&12/13/5&93/94/53 & 49.73\%\\
MT \textit{vs} \textsc{P-Soups} &9/13/8&7/11/12&7/10/13&7/8/15&7/4/19&8/10/12&10/9/11&10/13/7&65/78/97 & 45.45\%\\
\textsc{P-MORL} \textit{vs} \textsc{P-Soups} &16/9/5&7/15/8&9/11/10&8/2/20&14/7/9&10/7/13&11/13/6&11/3/16&86/67/87 & 56.21\%\\
\bottomrule        
\end{tabular}
\caption{Simulated pairwise win rate (\%) across all methods through Human Evaluation.} 
\label{table:human_eval_detailed2}
\end{table*} 

\section{Examples of \textsc{P-Soups} text generations}
\label{appen:examples}
Table \ref{table:empricial} shows empirical examples of the text generated from each preference combination of the 16 preference combination experiments for the same prompt. 
\begin{table}[h!]
\fontsize{7}{9}\selectfont
\centering
\begin{tabular}{ll}
\toprule
Prompt : ``Is a banana a fruit or a herb?" \\
\midrule
Unique Preference & Output Text \\ \midrule
AAA (Elementary, Concise, Friendly) & A banana is a fruit, not a herbal supplement. Just eat it and enjoy its deliciousness! Don't be a herhang goiter,\\ &  just eat the banana!\\ 
AAB (Elementary, Concise, Unfriendly) & A banana is a fruit and not a herbal plant. It grows on trees and when fully ripe, it needs to be harvested and eaten.\\ & Just like an apple, it is a type of plant that comes from trees.\\
AAC (Elementary, Concise, Sassy) & A banana is not a fruit nor a her. It is a plant! Duh! So, do not be fooled by its yellow color and sweet taste.\\ & It's like a trickster fruit! So there!\\
AAD (Elementary, Concise, Sarcastic) & A banana is a fruit, whether you like eating it like an apple or just simply want to eat the seeds like an herbal\\ & ingredient in your soup, it's a fruit! I don't mean to sound harsh or anything, but come on now! Are you really\\ & asking this question? Bananas are a fruit indeed!\\
ABA (Elementary, Informative, Friendly) & A banana is a fruit! It's a long, curved, creamy yellow food that comes in bunches. It' GROWS ON A TREE \\ & and ROUND AND DRY WHEN IT PICKED. We eat it in lots of things like a yoghurt or with honey or milk\\ & for a sweet treat. But a banana can also be sliced like a cake! They are DELICIOUS and they'll make your \\ & day that little bit sweeter. What you call a Bananas end is what you call its roots? No, really, trying it today then please!\\
ABB (Elementary, Informative, Unfriendly) & A banana is a fruit, not a herbal product. It is an edible fruit that is grown on a banana plant. One of the key \\ & characteristics of a fruit is that it contains seeds. Bananas do not have any seeds, so it is not a fruit in the traditional \\ & sense. However, they are a part of the herbaceous family, which suggests that they are related to herbs. This is why \\ & they are sometimes considered to be an herbal or medicinal product, even though they are ultimately a fruit.\\ & Could you please tell me if you have any more questions about this topic? I am happy to help you with any\\ & additional information you might require!\\
ABC (Elementary, Informative, Sassy) & A banana is a fruit! It's a tall, curved plant with a brown outer peel, a sweet white inside, and seeds in the middle.\\ & It grows on trees in tropical places like Hawaii and Brazil. Folks have enjoyed eating bananas for thousands of years. \\ & In fact, the ancient Greeks and Romans thought that bananas were so special that they used them as go-betweens in \\ & love letters! They'd stick a love message in a banerella (banana plus messenger) and send it across the sea. Sweet!\\ & So there you have it, bananas are like super-special fruits that have been making people fall in love for a very, very \\ & long time. So next time someone tells you that bananas aren't as classy as apples, just give 'em a big, juicy fruit\\ & and watch the smiles spread like wildfire!\\
ABD (Elementary, Informative, Sarcastic) & A banana is a fruit, not an herbal supplement. In fact, bananas are considered berries because they grow from a \\ & berry-type fruit group.\\
BAA (Expert, Concise, Friendly) & From a biological standpoint, a banana is classified as a fruit, as it develops from a flower and contains seeds. \\ & However, from a culinary perspective, bananas are often considered a herby vegetable, as they are used in savory\\ & dishes like curries or salads. So, I guess you could say it's a bit of both!\\
BAB (Expert, Concise, Unfriendly) & A banana is a fruit, not a herbal plant. Its consumption provides desired nutritional benefits, including vitamins \\ & and minerals, whereas herbal plants, such as sage or basil, are typically consumed for their medicinal or flavorful \\ & properties. If you want to discuss an academic paper related to this topic, I can suggest some relevant publications \\ & from a machine learning journal like "arXiv" or a natural language or cognitive science journal like the "Proceedings \\ & of the National Academy of Sciences" or the "Journal of the Royal Society, Interface"\\
BAC (Expert, Concise, Sassy) & A banana is a fruit, not a herbal plant. In botany, a fruit is often defined as a reproductive structure derived from the \\ & ovaries of a flowering plant, whereas a herby plant refers to a plant that is widely known for its medicinal and culinary \\ & properties. A banana, like all flowering plants, is a monocot, whereas most herbaceous plants, including mint and basil, \\ & are dicots. In conclusion, even if bananas shared a common monocotic base with some herbs, anatomical and evolutionary \\ & differences between the two plant types remain significant, and bananas still belong to the fruit category.\\
BAD (Expert, Concise, Sarcastic) & It's a fruit, but you can also use the stem to make tea. If you want, I can send you a research paper about the medicinal \\ & properties of banana stems.\\
BBA (Expert, Informative, Friendly) & From a plant biology perspective, a banana is a fruit. This is because bananas contain seeds and are typically developed \\ & from the flowering part of the plant. However, from an anatomical perspective, bananas can also be considered as a\\ & herbcaceous perennial. This means that they are herbaceous (they have no woody stems) and perennialize (they grow \\ & and produce for several years). So, in short, while bananas are typically classified as a fruit, they can also considered \\ & as an herb.\\
BBB (Expert, Informative, Unfriendly) & In a phytogenetic classification scheme, the banana is paraphyletic with respect to the genus Musa, \\ & whereupon the \"Musa\" clade would be recovered as the true monophyletic banana genus.  This would mean that the banana \\ & would be a herbcaceous perennial within the genus \"MUSA\", rather than a fruit.\\
BBC (Expert, Informative, Sassy) & A banana is a fruit, not a herbal plant. It is an edible fruits that belongs to the genus Musa of the family Musaceae. \\ & It's native origin is unknown, but it was first introduced to the Western world by Spanish explorers who found it growing \\ & in the coastal regions of northeastern South America. The banana we eat today is a domesticated form of the wild banana \\ & plant, known as Musa balbisiana. The domesticated banana wasn't selectively bred for its traits we see today. Instead, it \\ & resulted from a long process of random mutations by natural processes during the domestication process. The fruit of an adult \\ & banana tree will contain 23 nearly identical chromosomes, which all come from a single ancestral genome with multiple copies.\\ & This genome contains all the genes required for the development of the banana, including genes for resistance to viruses and \\ & fungi. Bananas are particularly susceptible to a fungus called black sigatoka, which attacks the flowers and a portion of the \\ & inner portion of each banana stalk. Black sigatokavirus can infect the plant, destroying the entire crop.\\
BBD (Expert, Informative, Sarcastic) & A banana is a fruit, not a herbal medicinal plant. However, in terms of culinary use, it is often treated like a herblmdal plant \\ & and used as an ingredient in many desserts and confections. If you're a fan of sweet treats, then you've probably partaken of \\ & banana flesh or stems coated in a creamy glaze or batter. You may not have known it at the time, but in that moment of delicious \\ & indulgence, you were consuming a fruit and a herbstalk.\\
\bottomrule
\end{tabular}
\caption{Example output generations for each of the 16 preference combinations from \textsc{P-soups}.}
\label{table:empricial}
\end{table}

\end{document}